\documentclass[final]{template}

\usepackage{times}
\usepackage{epsfig}
\usepackage{graphicx}
\usepackage{amsmath}
\usepackage{amssymb}
\usepackage{mathtools}
\usepackage{url}
\usepackage{color}
\usepackage{float}
\usepackage{array}
\usepackage{multirow}
\usepackage{arydshln}
\usepackage{longtable}

\usepackage{array}
\usepackage{multicol}
\usepackage{lipsum}
\usepackage{adjustbox}
\usepackage{enumitem}
\usepackage[pagebackref=true,breaklinks=true,colorlinks,bookmarks=false]{hyperref}

\begin{document}

\title{Lightweight Monocular Depth with a Novel Neural Architecture Search Method}

\author{Lam Huynh$^1$ \qquad
Phong Nguyen$^1$ \qquad
Jiri Matas$^2$ \\
Esa Rahtu$^3$ \qquad
Janne Heikkil\"a$^1$ \\
\small{$^1$Center for Machine Vision and Signal Analysis, University of Oulu} \\ 
\small{$^2$ Center for Machine Perception, Czech Technical University, Czech Republic} \\ 
\small{$^3$Computer Vision Group, Tampere University} }

\maketitle

\begin{abstract}

This paper presents a novel neural architecture search method, called LiDNAS, for generating lightweight monocular depth estimation models.
Unlike previous neural architecture search (NAS) approaches, where finding optimized networks is computationally highly demanding, the introduced  novel Assisted Tabu Search leads to efficient architecture exploration.
Moreover, we construct the search space on a pre-defined backbone network to balance layer diversity and search space size.
The LiDNAS method outperforms the state-of-the-art NAS approach, proposed for disparity and depth estimation, in terms of search efficiency and output model performance. 
The LiDNAS optimized models achieve result superior to compact depth estimation state-of-the-art on NYU-Depth-v2, KITTI, and ScanNet, while being $7\%$-$500\%$ more compact in size, i.e the number of model parameters.

\end{abstract}

\section{Introduction}

Depth information is essential to numerous computer vision applications, including robotics, mixed reality, and scene understanding. Traditionally, accurate depth measurements are acquired using stereo or multi-view setups~\cite{hartley2003multiple,szeliski2011structure} or active sensors such as ToF cameras, LIDARs. However, deploying such methods for resource-limited devices is costly or may even be infeasible in practice. Considering this, current advances in learning-based monocular depth estimation proffering them as viable alternatives to conventional approaches.

\begin{figure}[!t]
\begin{center}
  \includegraphics[width=0.84\linewidth]{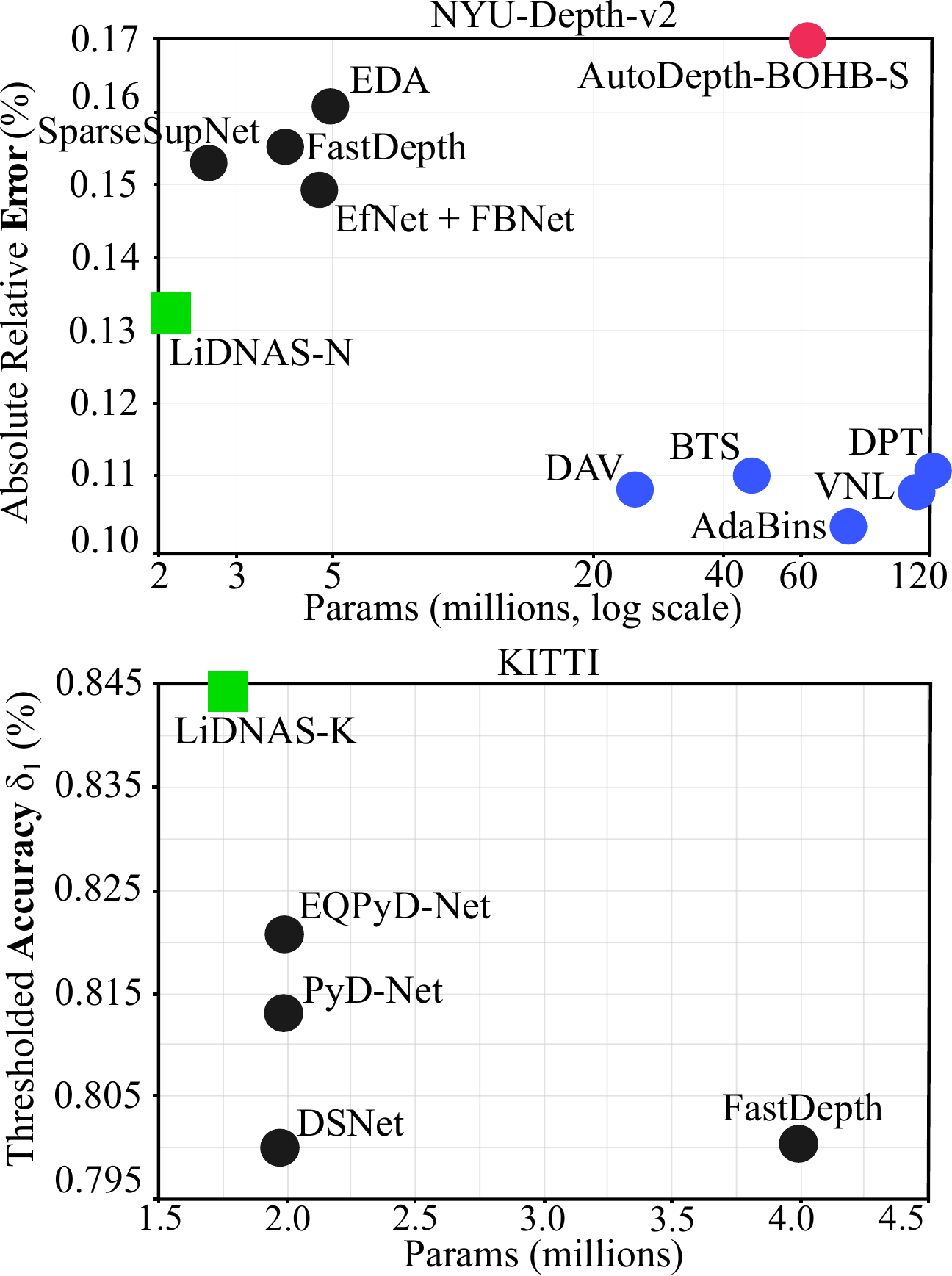}
\end{center}
  \caption{
  Absolute relative error and thresholded accuracy ($\delta_1$) vs.~the number of parameters for recent depth estimation methods on NYU-Depth-v2 (top) and KITTI (bottom) -- the LiDNAS models outperforms the lightweight baselines (black), while using substantially less parameters than the current state-of-the-art methods (blue). Compared to the recent NAS-based depth estimation method (red), LiDNAS improves in both performance and compactness.
  } \vspace{-0.45cm}
\label{fig:rel_params_chartv1} %
\end{figure}

Recent deep neural networks (DNN) show compelling results on single image depth estimation by formulating apparent depth cues~\cite{bhat2021adabins,chen2019structure,facil2019cam,Hu2018RevisitingSI,huynh2020guiding,lee2019big,liu2019planercnn,liu2018planenet,ramamonjisoa2019sharpnet,yang2021transformers} or estimating relative depth in unconstrained settings~\cite{chen2016single,lee2019monocular,Ranftl2021}. Moreover, self-supervised methods~\cite{garg2016unsupervised,godard2019digging} offered appealing solutions for single image depth estimation. Nevertheless, most studies focus on increasing accuracy at the expense of model complexity, making them infeasible to devices with limited hardware capabilities. 

To tackle this problem, lightweight depth estimation methods~\cite{wofk2019fastdepth} were proposed by utilizing small and straightforward architecture. Usually such simple designs are unreliable and yield low-quality predictions. Other popular strategies include quantizing the weights of a network into low-precision fixed-point operations~\cite{han2015deep} or pruning by directly cut off less important filters~\cite{yang2018netadapt}. That being said, these methods depend on a baseline model, tend to degrade its performance afterward and incapable of exploring new combinations of DNN operations. Moreover, creating a resource-constrained model is a non-trivial task requiring 1) expert knowledge to carefully balance accuracy and resource and 2) plenty of tedious trial-and-error work.

Neural architecture search (NAS), proposed recently~\cite{zoph2016neural,zoph2018learning}, exhibits compelling results, and more importantly, promises to relieve from the manual tweaking of deep neural architectures. Unfortunately, NAS methods mostly obligate thousands of training hours on hundreds of GPUs. To address this, recent NAS studies introduced various efficiency increasing techniques, which include weight sharing~\cite{pham2018efficient}, and network transformation~\cite{elsken2018efficient}. These methods show promising results, but they are still expensive and mainly focus on classification and detection. 

This paper introduces LiDNAS, an efficient model compactness-aware NAS framework, with the objective of searching for accurate and lightweight monocular depth estimation architectures. 
The approach is based on two main ideas. First, we observe that previous NAS methods essentially search for a few types of cells and then repeatedly accumulate the same cells to build the whole network. Although doing this simplifies the search process, it also restrains layer diversity that is important for computational efficiency. Instead, we construct a pre-defined backbone network that utilizes different layers striving for the right balance between flexibility and search space size. Secondly, we proposed the Assisted Tabu Search (ATS) for efficient neural architecture search. Inspired by the recent NAS study that suggests estimating network performance without training~\cite{mellor2021neural}, we integrate this idea into our multi-objective search function to swiftly evaluate our candidate networks. This, in turn, reduces $\sim 90\%$ search time compared to state-of-the-art NAS-based disparity and depth estimation  approaches~\cite{saikia2019autodispnet}. 

Figure~\ref{fig:rel_params_chartv1} summarizes a comparison between our LiDNAS models and
other state-of-the-art lightweight approaches. Compared to PyD-Net~\cite{poggi2018towards}, our method improves the REL, RMSE, and thresholded accuracy by $13.6\%$, $8.3\%$, and $3\%$ with similar execution time on the Google Pixel 3a phone (see Table~\ref{tab:runtime_comparison}). Compared to FastDepth~\cite{wofk2019fastdepth} and EDA~\cite{tu2020efficient}, our model achieves higher accuracy with fewer parameters.
To summarize, our work makes the following contributions: 

\begin{itemize}[noitemsep,topsep=3pt,parsep=3pt,partopsep=3pt]
\item We propose a multi-objective exploration framework, LiDNAS, searching for accurate and lightweight monocular depth estimation architectures.
\item We introduce a novel scheme called Assisted Tabu Search, enabling fast neural architecture search.
\item We create a well-defined search space that allows computational flexibility and layer diversity.
\item We achieve the state-of-the-art results compared to the lightweight baselines on NYU-Depth-v2, KITTI, and ScanNet while using less parameters.
\end{itemize}
\noindent The implementation of LiDNAS will be made publicly available upon publication of the paper.

\begin{figure*}[!t]
\begin{center}
  \includegraphics[width=0.99\linewidth]{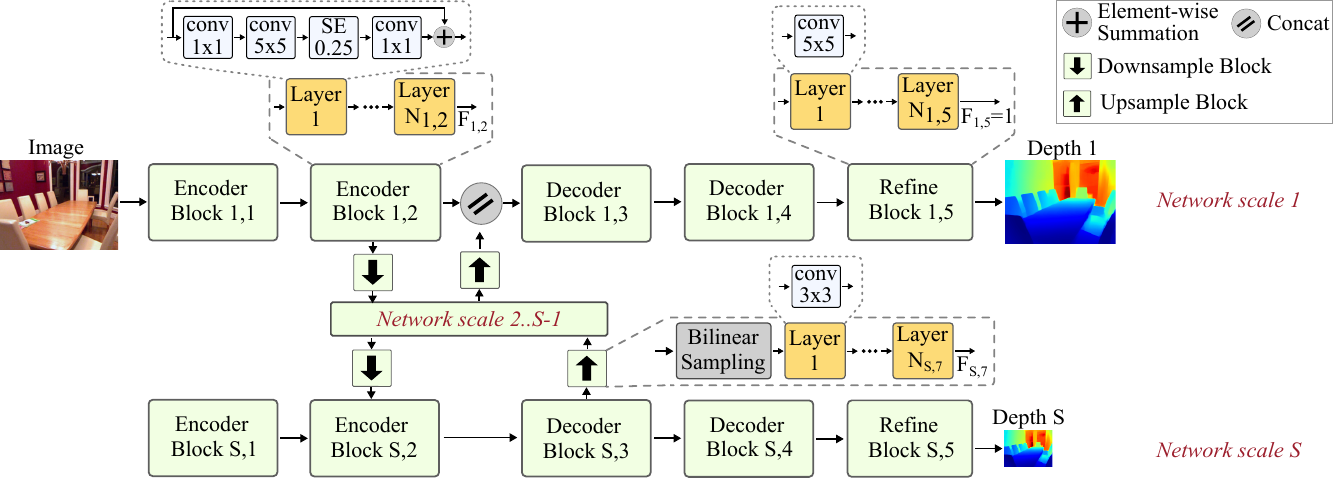}
\end{center}
  \caption{The search space of our LiDNAS framework.  Models are constructed from a pre-defined backbone network containing encoder, decoder, refine, downsample and upsample \textit{blocks} (green).  A block is formed by several identical layers (orange) that are generated from a pool of operations and connections. Layers within a block are the same while layers of different blocks can be different.} \vspace{-0.45cm}
\label{fig:search_space_ver2}
\end{figure*}

\begin{figure}[!b]
\begin{center}
\vspace{-0.45cm}
  \includegraphics[width=0.99\linewidth]{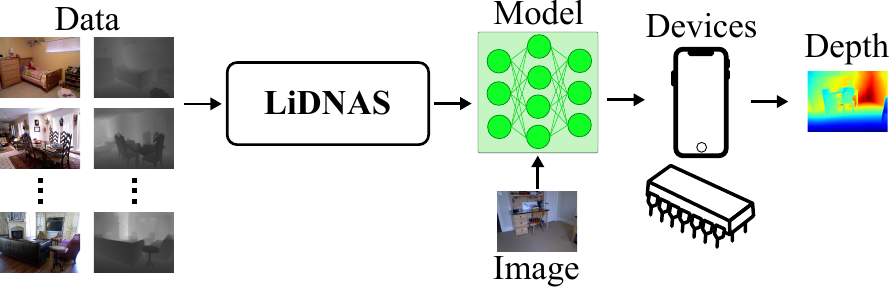}
\end{center}
  \caption{Overview of the proposed approach.}
\label{fig:overview}
\end{figure}

\section{Related work}
\noindent \textbf{Monocular depth estimation} 
Learning-based single image depth estimation was first introduced by Saxena et al.~\cite{saxena2006learning} and Eigen et al.~\cite{eigen2015predicting,eigen2014depth}. Later studies improved accuracy by using large network architectures~\cite{chen2019structure,Hu2018RevisitingSI,laina2016deeper} or integrating semantic information~\cite{jiao2018look} and surface normals~\cite{qi2018geonet}. Fu et al.~\cite{fu2018deep} formulated depth estimation as an ordinal regression problem, while~\cite{chen2016single,lee2019monocular} estimated relative instead of metric depth.  Facil et al.~\cite{facil2019cam} proposed to learn camera calibration from the images for depth estimation. Recent approaches further improve the performance by exploiting monocular priors such as planarity constraints~\cite{liu2019planercnn,liu2018planenet,Yin2019enforcing,huynh2020guiding,lee2019big} or occlusion~\cite{ramamonjisoa2019sharpnet}. Gonzalez and Kim~\cite{gonzalezbello2020forget} estimated depth by synthesizing stereo pairs from a single image, while~\cite{yang2021transformers} and~\cite{Ranftl2021} applied vision-transformer for depth prediction. However, these studies mostly focus on increasing accuracy at the cost of model complexity that is infeasible in resource-limited settings.

\noindent \textbf{Lightweight depth estimation architectures} For resource-limited hardware, it is more desirable to not only have a fast but also accurate model. One simple alternative is employing lightweight architectures such as MobileNet~\cite{howard2019searching,howard2017mobilenets,sandler2018mobilenetv2,wofk2019fastdepth}, GhostNet~\cite{han2020ghostnet}, and FBNet~\cite{tu2020efficient}. One popular approach is utilizing network compression techniques, including quantization~\cite{han2015deep}, network pruning~\cite{yang2018netadapt}, and knowledge distillation~\cite{yucel2021real,aleotti2021real}. Other methods employing well-known pyramid networks or dynamic optimization schemes. However, these tasks are tedious, require a lot of trial-and-error, and usually lead to architectures with low accuracy.

\noindent \textbf{Neural Architecture Search} There has been increasing interest in automating network design using neural architecture search. Most of these methods focus on searching high-performance architecture using reinforcement learning~\cite{baker2016designing,liu2018progressive,pham2018efficient,zoph2016neural,zoph2018learning}, evolutionary search~\cite{real2019regularized}, differentiable search~\cite{liu2018darts}, or other learning algorithms~\cite{luo2018neural}. However, these methods are usually very slow and require huge resources for training. Other studies~\cite{dong2018dpp,elsken2018multi,hsu2018monas} also attempt to optimize multiple objectives like model size and accuracy. Nevertheless, their search
process optimizes only on small tasks like CIFAR. In contrast,
our proposed method targets real-world data such as NYU, KITTI and ScanNet.

\section{LiDNAS}

We propose the LiDNAS framework to search for accurate and lightweight monocular depth estimation architectures. The overview the our approach is presented in Figure~\ref{fig:overview}.  It takes in a dataset as input to search for the best possible model.  This model can be deployed for depth estimation on hardware-limited devices. The first subsection defines the search space while the remaining two describe our multi-objective exploration and search algorithm.

\subsection{Search Space}

Previous neural architecture search (NAS) studies demonstrated the significance of designing a well-defined search space. A common choice of NAS is searching for a small set of complicated cells from a smaller dataset~\cite{zoph2018learning,liu2018progressive,real2019regularized}. These cells are later replicated to construct the entire architecture that hindered layer diversity and suffered from domain differences~\cite{tan2019mnasnet}. On the other hand, unlike classification tasks, dense prediction problems involve mapping a feature representation in the encoder to predictions at larger spatial resolution in the decoder. 

\begin{figure}[!t]
\begin{center}
  \includegraphics[width=0.99\linewidth]{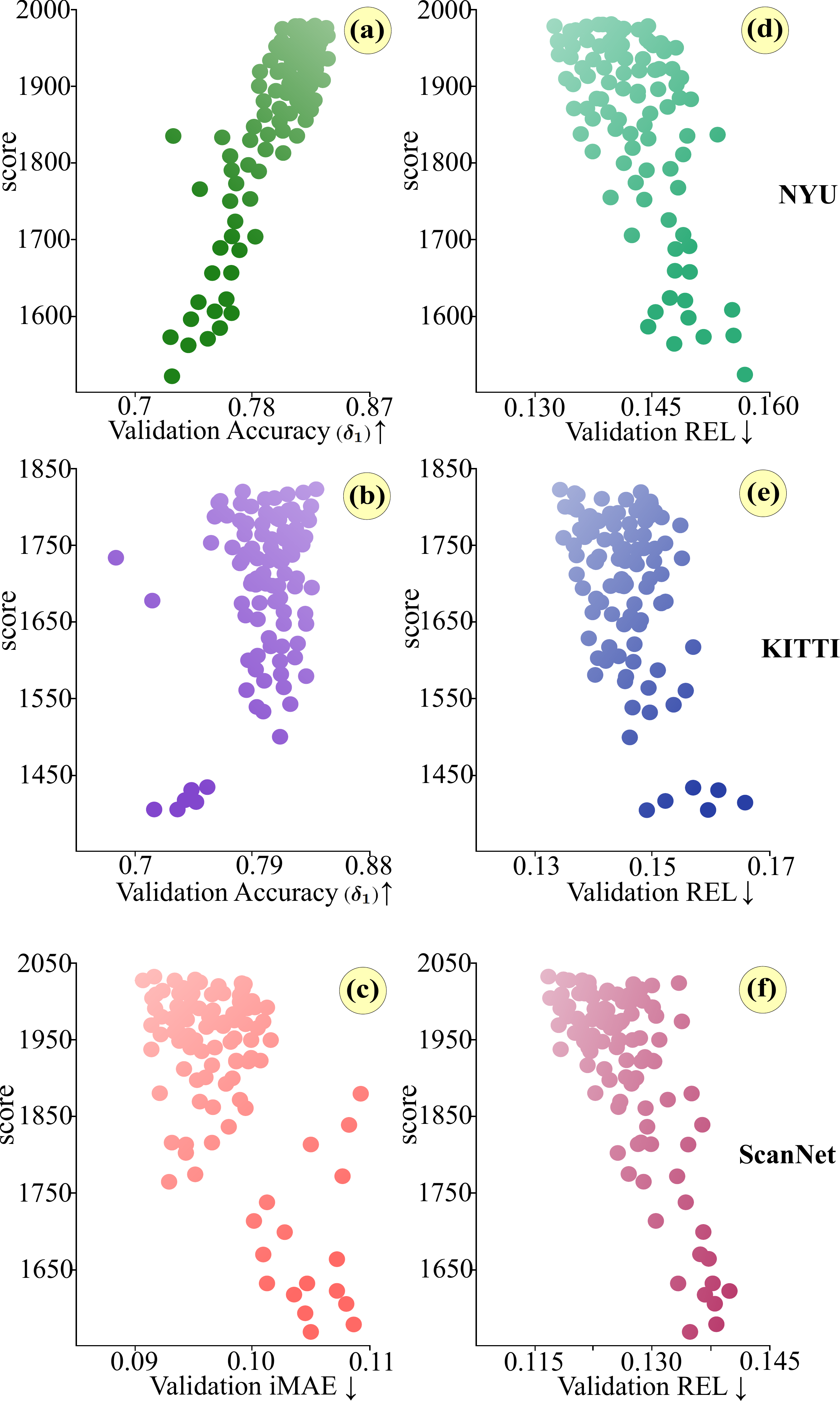}
\end{center}
  \caption{Plots of the \textit{score} at initialisation of untrained architectures against evaluation metrics after training: 
  (a), (b) accuracy ($\delta_{1}$);
  (e) mean absolute error of the inverse depth (iMAE);
  and (d), (e), (f) absolute relative error (REL). Plots from the first, second and third row are obtained from NYU-Depth-v2, KITTI and ScanNet dataset, respectively.} \vspace{-0.5cm}
\label{fig:naswot_scoring_ver2}
\end{figure}

To this end, we build our search space upon a pre-defined backbone that is shown as the set of green blocks in Figure~\ref{fig:search_space_ver2}. The backbone is divided into multi-scale pyramid networks operating at different spatial resolutions. Each network scale consists of two encoder blocks, two decoder blocks, a refine block, a downsample and a upsample block (except for scale 1). Each block is constructed from a set of identical layers (marked as orange in Figure~\ref{fig:search_space_ver2}). Inspired by~\cite{tan2019mnasnet}, we search for the layer from a pool of operations and connections, including:

\begin{figure*}[!t]
\begin{center}
  \includegraphics[width=0.99\linewidth]{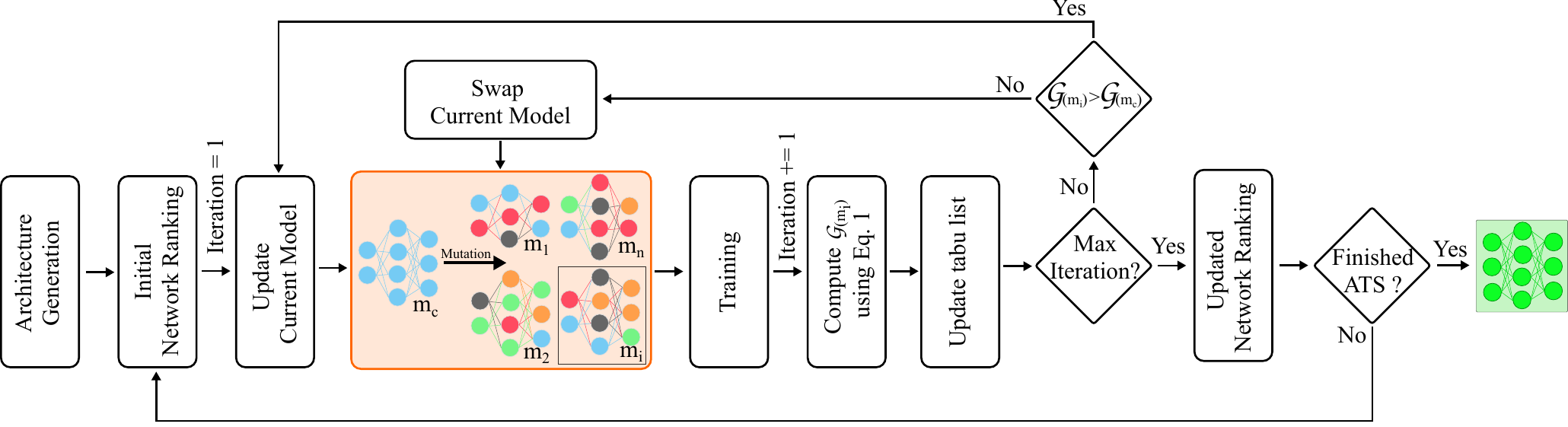}
\end{center}
  \caption{The flowchart of our architecture search that utilizes the Assisted Tabu Search (ATS) with mutation to search for accurate and lightweight monocular depth estimation networks.} \vspace{-0.45cm}
\label{fig:search_algorithm}
\end{figure*}

\begin{itemize}[noitemsep,topsep=0pt,parsep=1pt,partopsep=1pt]
    \item The number of resolution scales $S$.
    \item The number of layer for each block $N_{i,j}$.
    \item Convolutional operations (ConvOps): vanilla 2D convolution, depthwise convolution, and inverted bottleneck convolution.
    \item Convolutional kernel size (KSize): $3 \times 3$, $5 \times 5$.
    \item Squeeze and excitation ratio (SER): $0$, $0.25$.
    \item Skip connections (SOps): residual or no connection.
    \item The number of output channels: $F_{i,j}$.
\end{itemize}

\noindent where $i$ indicates the resolution scale and $j$ is the block index at the same resolution. Internal operations such as ConvOps, KSize, SER, SOps, $F_{i,j}$ are utilized to construct the layer while $N_{i,j}$ determines the number of layer that will be replicated for block$_{i,j}$. In other words, as shown in Figure~\ref{fig:search_space_ver2}, layers within a block (e.g. layers 1 to N$_{1,2}$ of Encoder Block 1,2 are the same) are similar while layers of different blocks (e.g. Layer 1 in Refine Block 1,5 versus Layer 1 in Upsample Block S,7) can be different.

We also perform layer mutation to further diversifying the network structure during the architecture search process. The mutation operations include:
\begin{itemize}[noitemsep,topsep=0pt,parsep=1pt,partopsep=1pt]
    \item Swapping operations of two random layers with compatibility check.
    \item Modifying a layer with a new valid layer from the predefined operations.
\end{itemize}

Moreover, we also set computational constraints to balance the kernel size with the number of output channels. Therefore, increasing the kernel size of one layer usually results in decreasing output channels of another layer. 

Assuming we have a network of $S$ scales, and each block has a sub-search space of size $M$ then our total search space will be $M^{5 + [(S - 1) * 7]}$. Supposedly, a standard case with $M=192$, $S=5$ will result in a search space of size $\sim 2 \times 10^{75}$.

\subsection{Multi-Objective Exploration}

We introduce a multi-objective search paradigm seeking for both accurate and compact architectures. For this purpose, we monitor the \textit{validation grade} $\mathcal{G}$ that formulates both accuracy and the number of parameter of a trained model. It is defined by

\begin{equation} 
\label{eq:validation_grade}
\mathcal{G}(m) = \alpha \times A(m) + (1 - \alpha) \times \bigg[\frac{P}{P(m)}\bigg]^{r}
\end{equation}

\noindent where $A(m)$ and $P(m)$ are validation accuracy and the number of parameters of model $m$. $P$ is the target compactness, $\alpha$ is the balance coefficient, and $r$ is an exponent with $r=0$ when $P(m) \leq P$ and otherwise $r=1$.  The goal is to search for an architecture $m^{*}$ where $G(m^{*})$ is maximum.

However, computing $G$ requires training for every architecture candidate, resulting in considerable search time. To mitigate this problem, Mellor et al.~\cite{mellor2021neural} suggested to score an architecture at initialisation to predict its performance before training. For a network $f$, the \textit{score(f)} is defined as:

\begin{equation} 
\label{eq:jacob_cov_score}
score(f) = log|K_{H}|
\end{equation}

\noindent where $K_{H}$ is the kernel matrix. 
Assume the mapping of model $f$ from a batch of data $X = \{x_i\}^{N}_{i=1}$ is $f(x_i)$. 
By assigning binary indicators to every activation units in $f$, a linear region $x_i$ of data point $i$ is represented by the binary code $c_i$. The kernel matrix $K_{H}$ is defined as:

\begin{equation} 
\label{eq:kernel_matrix}
K_{H} = \begin{pmatrix}
N_{A} - d_{H}(c_1, c_1) & \dots & N_{A} - d_{H}(c_1, c_N) \\
\vdots & \ddots & \vdots \\
N_{A} - d_{H}(c_N, c_1) & \dots & N_{A} - d_{H}(c_N, c_N)
\end{pmatrix}
\end{equation}

\noindent where $N_A$ is the number of activation units, and $d_{H}(c_i, c_j)$ is the Hamming distance between two binary codes. Inspired by this principle, we generate and train a set of different architectures on NYU, KITTI, and ScanNet. We evaluate the performance of these models and visualize the results against the \textit{score} that in our case is the mapping of depth values within image batches. Plots in Figure~\ref{fig:naswot_scoring_ver2} show that models with higher \textit{score} tend to yield better results. Leveraging this observation, we 1) utilize the \textit{score} in our initial network ranking, and 2) define the mutation exploration reward $\mathcal{R}$ as:

\begin{equation} 
\label{eq:mutation_exploration_reward}
\mathcal{R}(m_i,m_j) = \alpha \times \frac{score(m_j)}{score(m_i)} + (1 - \alpha) \times \bigg[\frac{P}{P(m_j)}\bigg]^{r}
\end{equation}

\noindent where $m_j$ is a child network that is mutated from $m_i$ architecture.

\subsection{Search Algorithm}

The flowchart of our architecture search is presented in Figure~\ref{fig:search_algorithm}. We first randomly generate $60K$ unique parent models and create the initial network ranking based on the \textit{score} in Eq.~\ref{eq:jacob_cov_score}. We then select \textit{six} architectures in which \textit{three} are the highest-ranked while the other \textit{three} have the highest score of the networks with the size closest to the target compactness. 

Starting from these initial networks, we strive for the best possible model utilizing the Assisted Tabu Search (ATS). Tabu search (TS)~\cite{glover1986future} is a high level procedure for solving multicriteria optimization problems. It is an iterative algorithm that starts from some initial feasible solutions and aims to determine better solutions while being designed to avoid traps at local minima. 

We propose ATS by applying Eq.~\ref{eq:validation_grade} and~\ref{eq:mutation_exploration_reward} to TS to speed up the searching process. Specifically, we mutate numerous children models ($m_1$, $m_2$, .., $m_n$) from the current architecture ($m_c$). The mutation exploration reward $\mathcal{R}(m_c, m_i)$ is calculated using Eq.~\ref{eq:mutation_exploration_reward}.
ATS then chooses to train the mutation with the highest rewards (e.g. architecture $m_i$ as demonstrated in Figure~\ref{fig:search_algorithm}). The validation grade of this model $\mathcal{G}(m_i)$ is calculated after the training. The performance of the chosen model is assessed by comparing $\mathcal{G}(m_i)$ with $\mathcal{G}(m_c)$. If $\mathcal{G}(m_i)$ is larger than $\mathcal{G}(m_c)$, then $m_i$ is a good mutation, and we opt to build the next generation upon its structure. Otherwise, we swap to use the best option in the tabu list for the next mutation. The process stops when reaching a maximum number of iterations or achieving a terminal condition. The network ranking will be updated, and the search will continue for the remaining parent architectures.

\begin{table*}[t!]
\caption{\label{tab:eval_nyuv2}Evaluation on the NYU-Depth-v2 dataset. Metrics with $\downarrow$ mean lower is better and $\uparrow$ mean higher is better. Type column shows the exploration method used to obtain the model. RL, ATS, and manual, refer to reinforcement learning, assisted tabu search, and manual design, respectively.}
\centering
\small
\begin{tabular}{@{}llrcccccccc@{}}
\hline
\multicolumn{2}{c}{\textbf{Architecture}} & \textbf{\#params} & \textbf{Type} & \textbf{Search Time} & \textbf{REL$\downarrow$} & \textbf{RMSE$\downarrow$} & \(\boldsymbol{\delta_{1}}\)$\uparrow$ & \(\boldsymbol{\delta_{2}}\)$\uparrow$ & \(\boldsymbol{\delta_{3}}\)$\uparrow$ \\ \hline

AutoDepth-BOHB-S & Saikia et al.'19~\cite{saikia2019autodispnet} & 63.0M & RL & 42 GPU days & 0.170 & 0.599 & - & - & - \\ \hline

EDA & Tu et al.'21~\cite{tu2020efficient} & 5.0M & Manual & - & 0.161 & 0.557 & 0.782 & 0.945 & 0.984 \\ \hline

FastDepth & Wofk et al.'19~\cite{wofk2019fastdepth} & 3.9M & Manual & - & 0.155 & 0.599 & 0.778 & 0.944 & 0.981 \\ \hline

SparseSupNet & Yucel et al.'21~\cite{yucel2021real} & 2.6M & Manual & - & 0.153 & 0.561 & 0.792 & 0.949 & 0.985 \\ \hline

Ef+FBNet & Tu \& Wu et al.~\cite{tu2020efficient,wu2019fbnet}  & 4.7M & Manual & - & 0.149 & 0.531 & 0.803 & 0.952 & 0.987 \\ \hline

LiDNAS-N & Ours  & \textbf{2.1M} & ATS & 4.3 GPU days & \textbf{0.132} & \textbf{0.487} & \textbf{0.845} & \textbf{0.965} & \textbf{0.993} \\ \hline

\end{tabular} \vspace{-0.45cm}
\end{table*}

\section{Performance analysis}

In this section, we evaluate the performance of the proposed method and compare it with several baselines on the NYU-Depth-v2, KITTI, and ScanNet datasets. 

\subsection{Datasets} We evaluate the proposed method using NYU-Depth-v2 \cite{silberman2012indoor}, ScanNet~\cite{dai2017scannet} and KITTI~\cite{geiger2013vision} datasets. NYU-Depth-v2 contains $\sim120K$ RGB-D images obtained from 464 indoor scenes. From the entire dataset, we use 50K images for training and the official test set of 654 images for evaluation. The ScanNet dataset comprises of 2.5 million RGB-D images acquired from 1517 scenes. For this dataset, we use the training subset of $\sim20K$ images provided by the Robust Vision Challenge 2018 \cite{robustvision2018} (ROB). In this paper, we report the results on the ScanNet official test set of $5310$ images instead. KITTI is an outdoor driving dataset, where we use the standard Eigen split~\cite{eigen2015predicting,eigen2014depth} for training (39K images) and testing (697 images).

\subsection{Evaluation metrics} The performance is assessed using the standard metrics provided for each dataset. That is, for NYU-Depth-v2 and KITTI we calculate the mean absolute relative error (REL), root mean square error (RMSE), and thresholded accuracy ($\delta_i$). For the ScanNet dataset, we provide the mean absolute relative error (REL), mean square relative error (sqREL), scale-invariant mean squared error (SI), mean absolute error (iMAE), and root mean square error (iRMSE) of the inverse depth values. 

\subsection{Implementation Details} \label{implementation_detail}

For searching, we directly perform our architecture exploration on the training samples of the target dataset. We set the target compactness parameter $P$ using the previously published compact models as a guideline. We set the maximum number of exploration iteration to 100 and stop the exploration procedure if a better solution cannot be found after 10 iterations. The total search time required to find optimal architecture is $\sim 4.3$ GPU days.

 For training, we use the Adam optimizer~\cite{kingma2014adam} with $(\beta_1, \beta_2, \epsilon) = (0.9, 0.999, 10^{-8})$. The initial learning rate is $7*10^{-4}$, but from epoch 10 the learning is reduced by $5\%$ per $5$ epochs. We use batch size 256 and augment the input RGB and ground truth depth images using random rotation ([-5.0, +5.0] degrees), horizontal flip, rectangular window droppings, and colorization (RGB only).

\subsection{Comparison with state-of-the-art}
\vspace{-5mm}
\noindent \paragraph{NYU-Depth-v2:} 

\begin{figure*}[!t]
\begin{center}
  \includegraphics[width=0.935\linewidth]{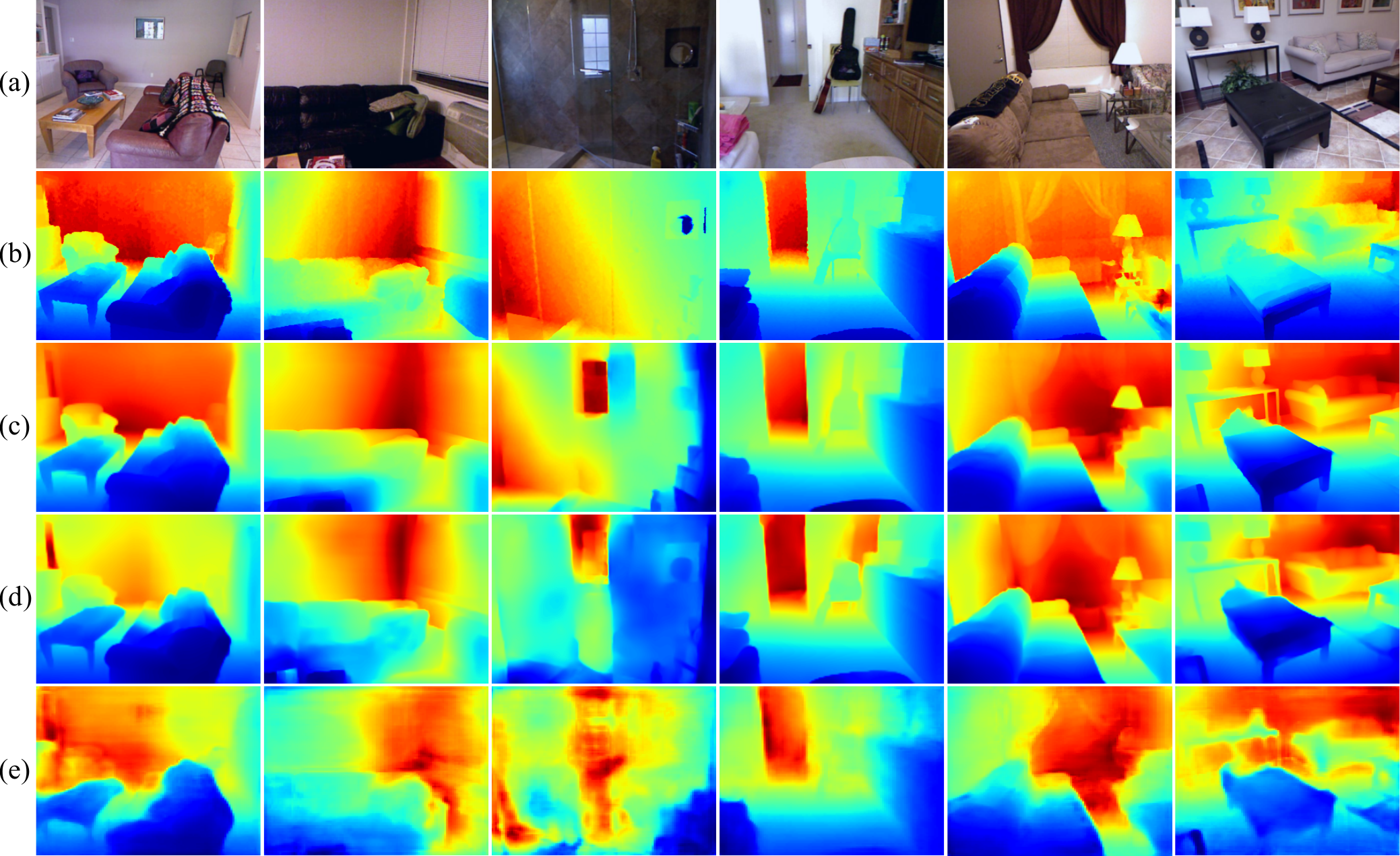}
\end{center}
  \caption{Comparison on the NYU test set. 
  (a) input image,
  (b) ground truth,
  (c) LiDNAS-N,
  (d) Ef+FBNet~\cite{tu2020efficient,wu2019fbnet},
  and (e) FastDepth~\cite{wofk2019fastdepth}.}
\label{fig:qualitative_nyu}
\end{figure*}

We set the target compactness $P=2M$ with the balance coefficient $\alpha=0.6$ to search for the optimized model on NYU-Depth-v2. We then select the best performance model (LiDNAS-N) and compare its results with lightweight state-of-the-art methods~\cite{tu2020efficient,wofk2019fastdepth,wu2019fbnet,yucel2021real} along with their numbers of parameters. As shown in Table~\ref{tab:eval_nyuv2}, LiDNAS-N outperforms the baseline while containing the least amount of parameters. Comparing with the best-performing approach, the proposed model improves the REL, RMSE, and $\theta_{1}$ by $11.4\%$, $8.2\%$, and $6.8\%$ while compressing the model size by $55\%$. Our method produces high-quality depth maps with sharper details as presented in Figure~\ref{fig:qualitative_nyu}. However, we observe that all methods still struggle in challenging cases, such as the scene containing Lambertian surfaces as illustrated by the example in the third column of Figure~\ref{fig:qualitative_nyu}. Moreover, the proposed method improves REL and RMSE by $22.3\%$ and $18.7\%$ while using only $3\%$ of the model parameters comparing to the state-of-the-art NAS-based disparity and depth estimation approaches~\cite{saikia2019autodispnet}. In addition, our method requires $90\%$ less search time than~\cite{saikia2019autodispnet}.

\begin{table}[b!]
\caption{\label{tab:eval_kitti}Evaluation on the KITTI dataset. Metrics with $\downarrow$ mean lower is better and $\uparrow$ mean higher is better.}
\centering
\small
\adjustbox{width=\columnwidth}{\begin{tabular}{@{}lrccccc@{}}
\hline
\textbf{Method} &\textbf{\#params} & \textbf{REL$\downarrow$} & \textbf{RMSE$\downarrow$} & \(\boldsymbol{\delta_{1}}\)$\uparrow$ & \(\boldsymbol{\delta_{2}}\)$\uparrow$ & \(\boldsymbol{\delta_{3}}\)$\uparrow$ \\ \hline

FastDepth~\cite{wofk2019fastdepth} & 3.93M & 0.156 & 5.628 & 0.801 & 0.930 & 0.971 \\ \hline

PyD-Net~\cite{poggi2018towards} & 1.97M & 0.154 & 5.556 & 0.812 & 0.932 & 0.970 \\ \hline

EQPyD-Net~\cite{cipolletta2021energy} & 1.97M & 0.135 & 5.505 & 0.821 & 0.933 & 0.970 \\ \hline

DSNet~\cite{aleotti2021real} & 1.91M & 0.159 & 5.593 & 0.800 & 0.932 & 0.971 \\ \hline

LiDNAS-K & \textbf{1.78M} & \textbf{0.133} & \textbf{5.157} & \textbf{0.842} & \textbf{0.948} & \textbf{0.980} \\ \hline

\end{tabular}} %
\end{table}

\begin{figure*}[!t]
\begin{center}
  \includegraphics[width=0.99\linewidth]{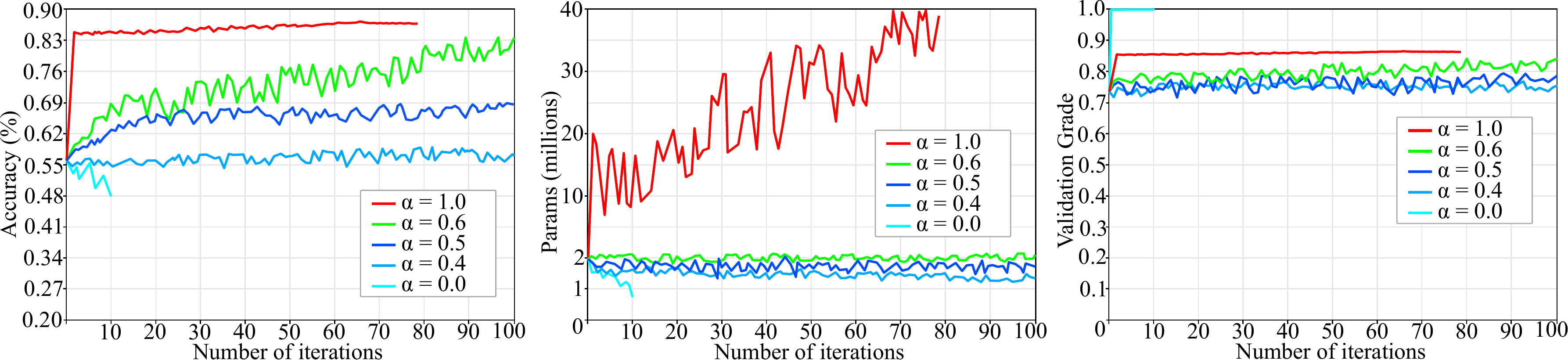}
\end{center}
  \caption{The progress of different searching scenarios on the NYU-Depth-v2 dataset. From left to right, charts show the accuracy, the number of parameters, and validation grade vs. the number of searching iterations.} \vspace{-0.25cm}
\label{fig:convergence_v2}
\end{figure*}

\vspace{-2mm}
\noindent \paragraph{KITTI:}

\begin{figure}[!b]
\begin{center}
\vspace{-2mm}
  \includegraphics[width=0.87\linewidth]{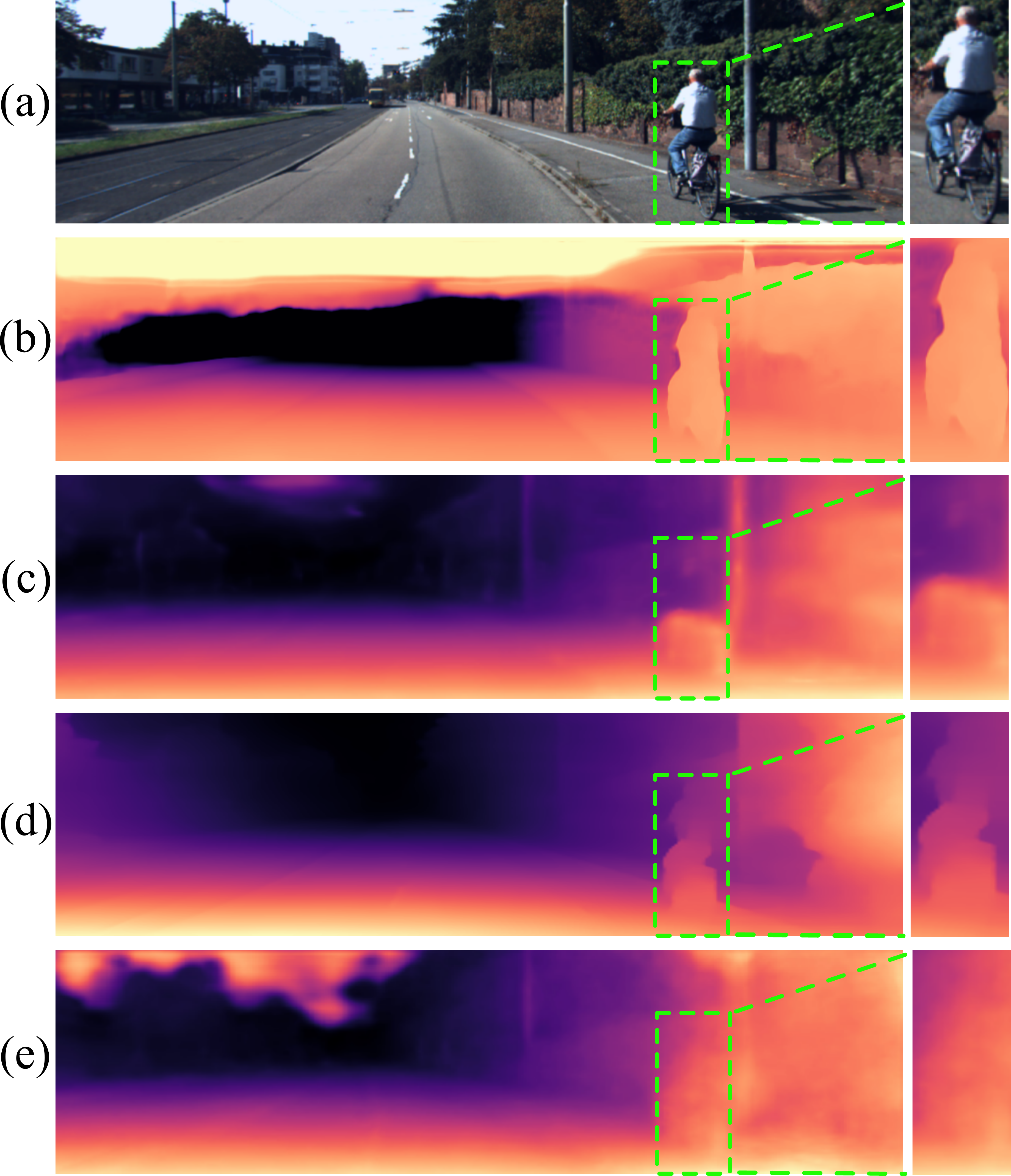}
\end{center}
  \caption{Comparison on the Eigen split of KITTI. 
  (a) input image,
  (b) LiDNAS-K,
  (c) DSNet~\cite{aleotti2021real},
  (d) PyD-Net~\cite{poggi2018towards},
  and (e) FastDepth~\cite{wofk2019fastdepth}.
  Images in the right column presented zoom-in view for better visualization.}
\label{fig:qualitative_kitti}
\end{figure}

In the case of KITTI, we aim at the target compactness of $P=1.5M$ with $\alpha=0.55$.  We then train our candidate architectures with the same self-supervised procedure proposed by~\cite{godard2019digging} and adopted by the state-of-the-art approaches~\cite{aleotti2021real,cipolletta2021energy,poggi2018towards,wofk2019fastdepth}. After the search, we pick the best architecture (LiDNAS-K) to compare with the baselines and report the performance figures in Table~\ref{tab:eval_kitti}. The LiDNAS-K model yields competitive results with the baselines while also being the smallest model. We observe that our proposed method provides noticeable improvement from PyD-Net and EQPyD-Net. Examples from Figure~\ref{fig:qualitative_kitti} show that the predicted depth maps from LiDNAS-K are more accurate and contain fewer artifacts.

\noindent \paragraph{ScanNet:}

For ScanNet, we set the target compactness to $4.5M$ with $\alpha=0.57$ for searching. Despite of being compact, our best performance model (LiDNAS-S) produces competitive results compared with state-of-the-art methods, as shown in Table~\ref{tab:eval_scannet}. More specifically, it requires only $20\%$ of the number of parameters in comparison with the best performance baseline. 
We also observe that although SARPN~\cite{chen2019structure} and Hu et al.~\cite{Hu2018RevisitingSI} models are multiple times larger than DS-SIDENet~\cite{ren2019deep} or DAV~\cite{huynh2020guiding}, the latter still yield better results, emphasizing the importance of optimal network structure. Furthermore, our model produces comparable depth maps as shown in Figure~\ref{fig:qualitative_scannet}. Details of the generated architectures are provided in the supplementary material.

\begin{table}[b!]
\caption{\label{tab:eval_scannet}Evaluation results on ScanNet \cite{dai2017scannet} dataset.}
\centering
\small
\adjustbox{width=\columnwidth}{\begin{tabular}{@{}lrccccc@{}}
\hline
\textbf{Architecture} & \textbf{\#params} & \textbf{REL} & \textbf{sqREL} & \textbf{SI} & \textbf{iMAE} & \textbf{iRMSE} \\ \hline

SARPN~\cite{chen2019structure} & 210.3M & 0.134 & 0.077 & \textbf{0.015} & 0.093 & 0.100 \\ \hline

Hu et al.~\cite{Hu2018RevisitingSI} & 157.0M & 0.139 & 0.081 & 0.016 & 0.100 & 0.105 \\ \hline

DS-SIDENet~\cite{ren2019deep} & 49.8M & 0.133 & \textbf{0.057} & - & - & - \\ \hline

DAV~\cite{huynh2020guiding} & 25.1M & 0.118 & \textbf{0.057} & \textbf{0.015} & \textbf{0.089} & \textbf{0.097} \\ \hline

LiDNAS-S & \textbf{5.2M} & \textbf{0.117} & 0.059 & \textbf{0.015} & 0.090 & \textbf{0.097} \\ \hline

\end{tabular}}
\end{table}

\vspace{-2mm}
\noindent \paragraph{Runtime Measurement:}

We also compare the runtime of our models with state-of-the-art lightweight methods on an Android device using the app from the Mobile AI benchmark developed by Ignatov et al.~\cite{ignatov2021fast}. To this end, we utilize the pre-trained models provided by the authors (Tensorflow~\cite{poggi2018towards}, PyTorch~\cite{wofk2019fastdepth}) and convert them to \textit{tflite}. Unfortunately, we can only report the measurement on the mobile CPU due to the technical issues occurring when converting PyTorch models to TFLite GPU delegate. That being said, the results in Table~\ref{tab:runtime_comparison} suggest that the proposed approaches produce competing performance, with the potential of running real-time on mobile devices with further optimization.

\begin{table}[!b]
\caption{\label{tab:runtime_comparison}Average runtime comparison of the proposed method and other lightweight models. Runtime values are measured using a Pixel 3a phone with input image resolution ($640 \times 480$).}
\centering
\small
\begin{tabular}{lc}
\hline
\textbf{Architecture} & \textbf{CPU(ms)} \\ \hline

FastDepth~\cite{wofk2019fastdepth} & 458 \\ \hline

Ef+FBNet~\cite{tu2020efficient,wu2019fbnet} & 852 \\ \hline

PyD-Net~\cite{poggi2018towards} & 226 \\ \hline

LiDNAS-K & \textbf{205} \\ \hline

LiDNAS-N & 262 \\ \hline

LiDNAS-S & 380 \\ \hline

\end{tabular}
\end{table}

\subsection{Ablation studies}
\noindent \paragraph{Exploration Convergence:}

We experiment with various settings for the multi-objective balance coefficient ($\alpha$) to assess its effect on the performance. For this purpose, we perform the architecture search with $\alpha$ set to $0.0$, $0.4$, $0.5$, $0.6$, and $1.0$ while the target compactness $P=2.0M$.  Figure~\ref{fig:convergence_v2} presents the searching progress for accuracy (left), the number of parameters (center), and validation grade (right) from one parent architecture on NYU-Depth-v2. We observe that, scenario with $\alpha=0.0$ quickly becomes saturated as it only gives reward to the smallest model. Searching with $\alpha=0.4$ favors models with compact size but also with limited accuracy. The case with $\alpha=0.5$ provides a more balanced option, but accuracy is hindered due to fluctuation during searching. The exploration with $\alpha=1.0$ seeks for the network with the best accuracy yet producing significantly larger architecture while the case where $\alpha=0.6$ achieves promising accuracy although with slightly bigger model than the target compactness.
\vspace{-2mm}
\noindent \paragraph{Searching Scenarios:}

\begin{figure}[!b]
\begin{center}
\vspace{-0.35cm}
  \includegraphics[width=0.65\linewidth]{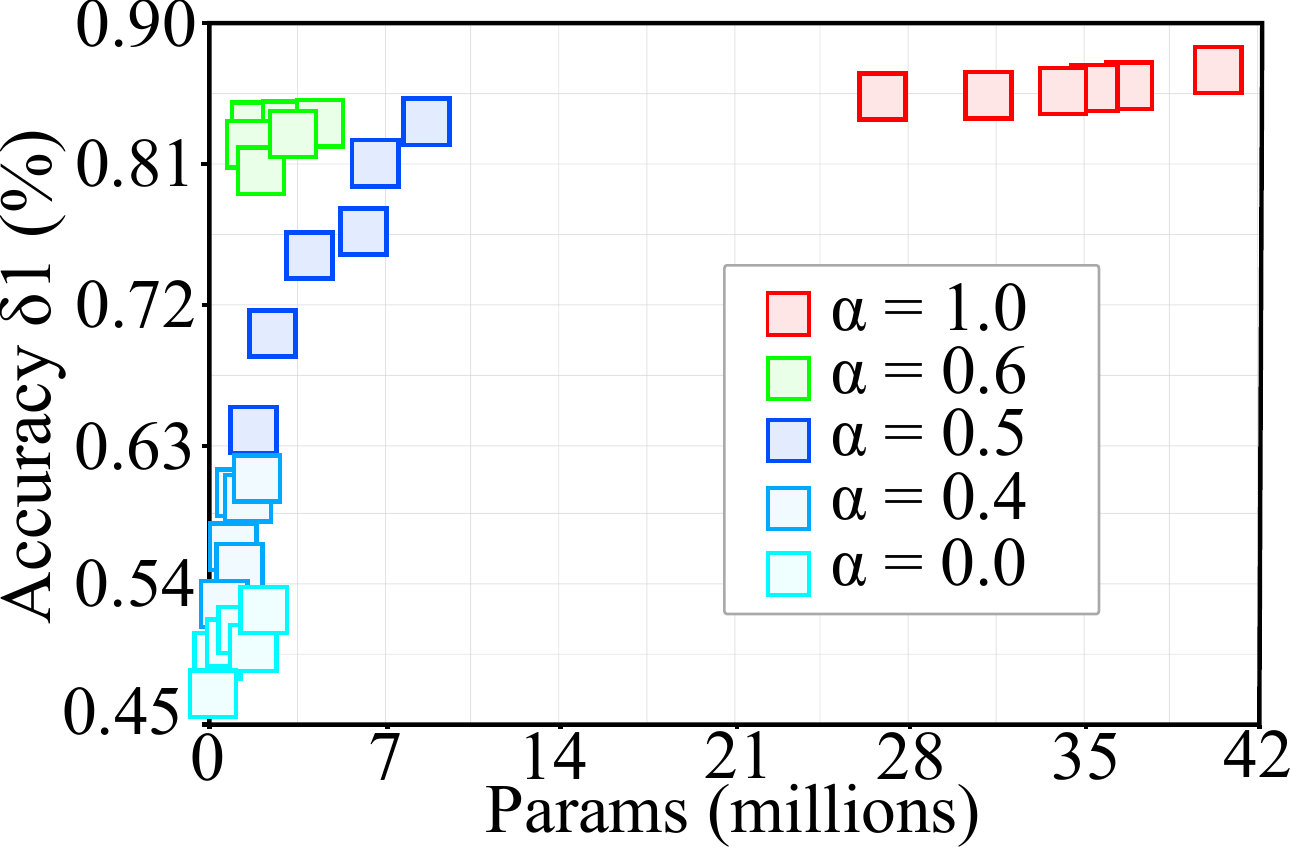}
\end{center}
\vspace{-0.35cm}
  \caption{Trade-off between accuracy vs. the number of parameters of best models trained with different searching scenarios on NYU-Depth-v2 dataset.}
\label{fig:searching_scenarios}
\end{figure}

To further analyze the outcome of different searching scenarios, we perform architecture searches for \textit{six} parent networks in five settings with $\alpha=0.0, 0.4, 0.5, 0.6, 1.0$ and $P=2.0M$ on NYU-Depth-v2. Results in Figure~\ref{fig:searching_scenarios} show that best performance models in case $\alpha=0.5$ are more spread out, while training instances with $\alpha=0.6$ tend to produce both accurate and lightweight architectures. This, in turn, emphasizes the trade-off between validation accuracy and network size.

\begin{figure*}[!t]
\begin{center}
  \includegraphics[width=0.935\linewidth]{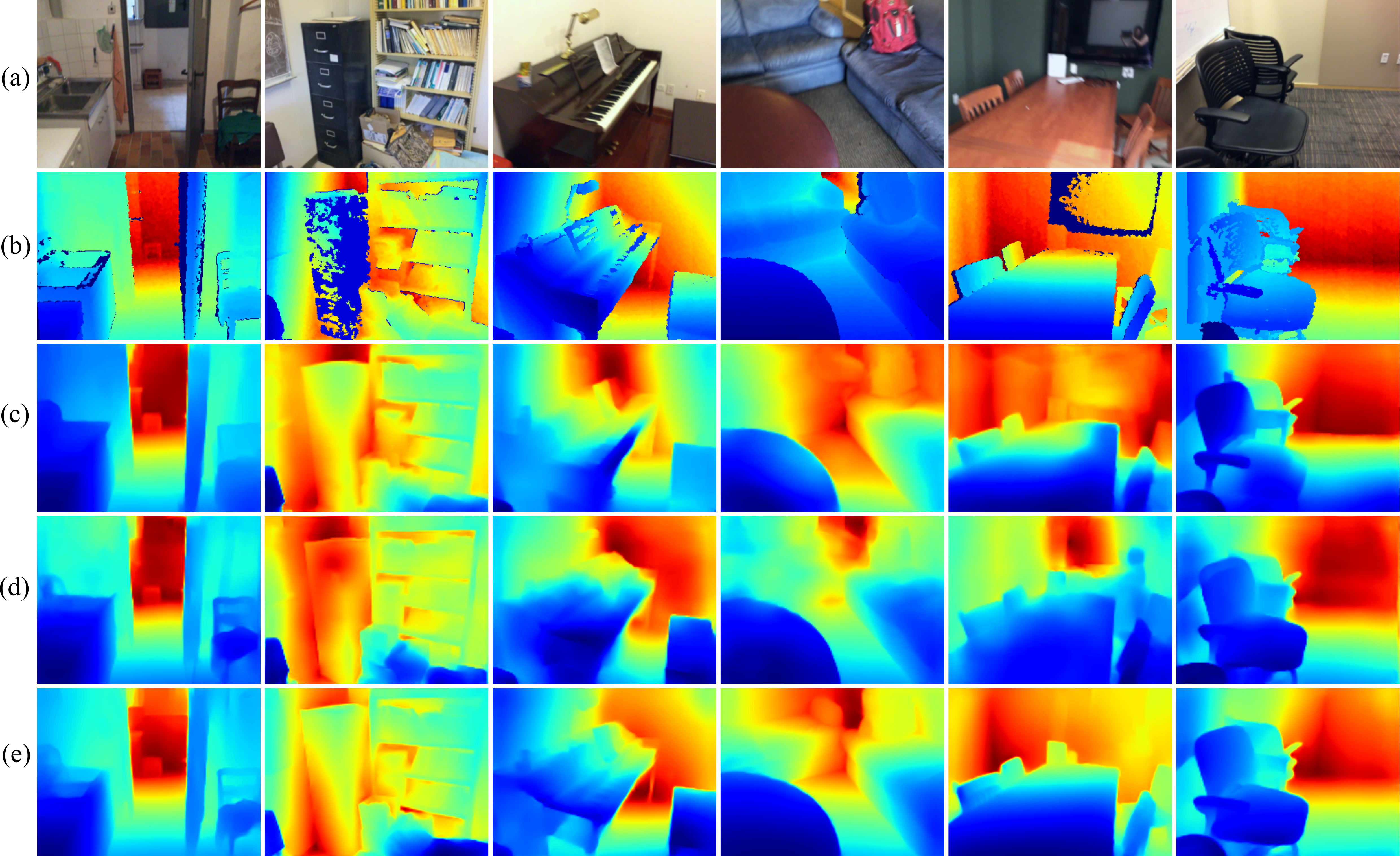}
\end{center}
  \caption{Comparison on the ScanNet test set. 
  (a) input image,
  (b) ground truth,
  (c) LiDNAS-S,
  (d) DAV~\cite{huynh2020guiding},
  and (e) SARPN~\cite{chen2019structure}.}
\label{fig:qualitative_scannet}
\end{figure*}

\section{Conclusion}

This paper proposed a novel NAS framework to construct lightweight monocular depth estimation architectures using Assisted Tabu Search and employing a well-defined search space for balancing layer diversity and search volume. The proposed method achieves competitive accuracy on three benchmark datasets while running faster on mobile devices and being more compact than state-of-the-art handcrafted and automatically generated models. Our work provides a potential approach towards optimizing the accuracy and the network size for dense depth estimation without the need for manual tweaking of deep neural architectures.

{\small
\bibliographystyle{ieee_fullname}
\bibliography{ms}
}

\end{document}